# Efficient coding along the visual hierarchy

Ananya Passi[1] , Brian S. Robinson[2], and Michael F. Bonner[1]

[1]Department of Cognitive Science, Johns Hopkins University, Baltimore, MD, USA
[2]Applied Physics Laboratory, Johns Hopkins University, Baltimore, MD, USA

[1]{apassi1, mfbonner}@jh.edu
[2]{brian.robinson}@jhuapl.edu

## ABSTRACT

Biological visual systems learn from limited experience, unlike deep learning models that rely on millions of training images. What learning principles make this possible? We tested whether efficient coding—the idea that neural representations capture the statistical structure of natural inputs—can build a hierarchy of human-aligned visual features from limited data. We developed an unsupervised learning procedure in which each layer of a deep network compresses its inputs onto the dominant modes of variation in natural images, using only local statistics and no labels, tasks, or backpropagation. This unsupervised procedure yields features that progress from edges and colors to textures and shapes. The features of this deep efficient coding model are readily recognized by human observers and are predictive of image-evoked fMRI responses in human visual cortex. Furthermore, a hybrid learning procedure that combines efficient coding with supervised fine-tuning yields better brain alignment in low-data settings and more rapid category learning. These findings suggest that efficient coding may shape representations across the entire visual hierarchy and help explain the data efficiency of biological vision.

## INTRODUCTION

Deep learning models achieve impressive performance in predicting visual cortex representations and visual behaviors (Khaligh-Razavi & Kriegeskorte, 2014; Kubilius et al., 2016; Rajalingham et al., 2018; Yamins et al., 2014). However, they fail to capture a key feature of biological vision: the ability to learn useful representations from little data (Lake et al., 2017; Zador, 2019). Instead, the prevailing paradigm of modern deep learning involves supervised or self-supervised updates over millions to billions of training samples. This discrepancy suggests that the representations of biological vision may be shaped by fundamentally different learning principles.

What principles might underlie the efficiency of visual learning in brains? One candidate mechanism is unsupervised learning, where neural systems adapt to the latent statistical structure of sensory stimuli without any task or explicit labels (Barlow, 1961; Bell & Sejnowski, 1997; Olshausen & Field, 1996, 2004; Zhong et al., 2025). Theoretical work has proposed a class of unsupervised learning principles in biology known as efficient coding (Barlow, 1961; Hermundstad et al., 2014; Olshausen & Field, 2004). Efficient coding theories propose that neural representations are shaped to convey as much information as possible about natural stimuli given the biological constraints under which the nervous system operates.

These theories have largely been applied to the representations of low-level visual processing, such as the retina and V1, with many implementations focusing on the objective of reducing redundancies in neural activity by making activations sparse (Atick & Redlich, 1992; Doi et al., 2012; Olshausen & Field, 1996, 1997; Vinje & Gallant, 2000). However, it has been hypothesized that efficient coding principles may have a role in shaping representations even at higher levels of visual processing (Beyeler et al., 2019; Karklin & Lewicki, 2003), with some work arguing for a different objective other than sparsity. One perspective proposes that in central visual processing, efficient codes are those that devote more resources to the most variable (and thus least predictable) features of natural sensory signals (Hermundstad et al., 2014). However, no previous studies have tested whether this efficient coding principle can be used to learn a deep hierarchy of human-aligned visual features.

In this work, we develop a hybrid approach to visual learning that combines unsupervised efficient coding with supervised learning, and we evaluate its ability to yield interpretable and task-relevant representations under the constraint of limited training data. Using a layer-wise efficient coding approach, our network first learns a hierarchy of visual features from natural images without labels, tasks, or backpropagation. This unsupervised stage alone produces behaviorally aligned features of increasing complexity along the hierarchy. When combined with supervised fine-tuning, the hybrid model strongly predicts human visual cortex representations after exposure to just 1,000 images. Furthermore, we show that this hybrid approach accelerates category learning, converging faster, achieving higher accuracy with fewer labeled examples, and maintaining performance even when early layers are frozen during fine-tuning. Together, these findings suggest that efficient coding principles may shape representations along the entire visual hierarchy and may be crucial to understanding the data efficiency of biological vision.

## RESULTS

### Deep efficient coding

We first sought to learn a hierarchy of visual representation based only on the statistical regularities of the sensory inputs and without a downstream task objective. To accomplish this, we developed an unsupervised learning procedure to capture the principal modes of variance for natural images in a deep network without any task-specific signal. We applied this procedure to a learned scattering network architecture, which separates fixed spatial filtering from learned channel mixing: each layer first applies a fixed set of wavelet filters followed by a learned 1×1 convolution that mixes information across channels (Guth et al., 2024). This combination of spatial filtering plus 1×1 convolution is functionally similar to the typical convolution operation applied in conventional convolutional networks, but it factorizes the weights over space versus channels. This has the key benefit of allowing for a strong inductive bias in which the spatial weights are fixed to be oriented bandpass filters and the learning problem is constrained to a linear projection applied to the outputs of these filters.

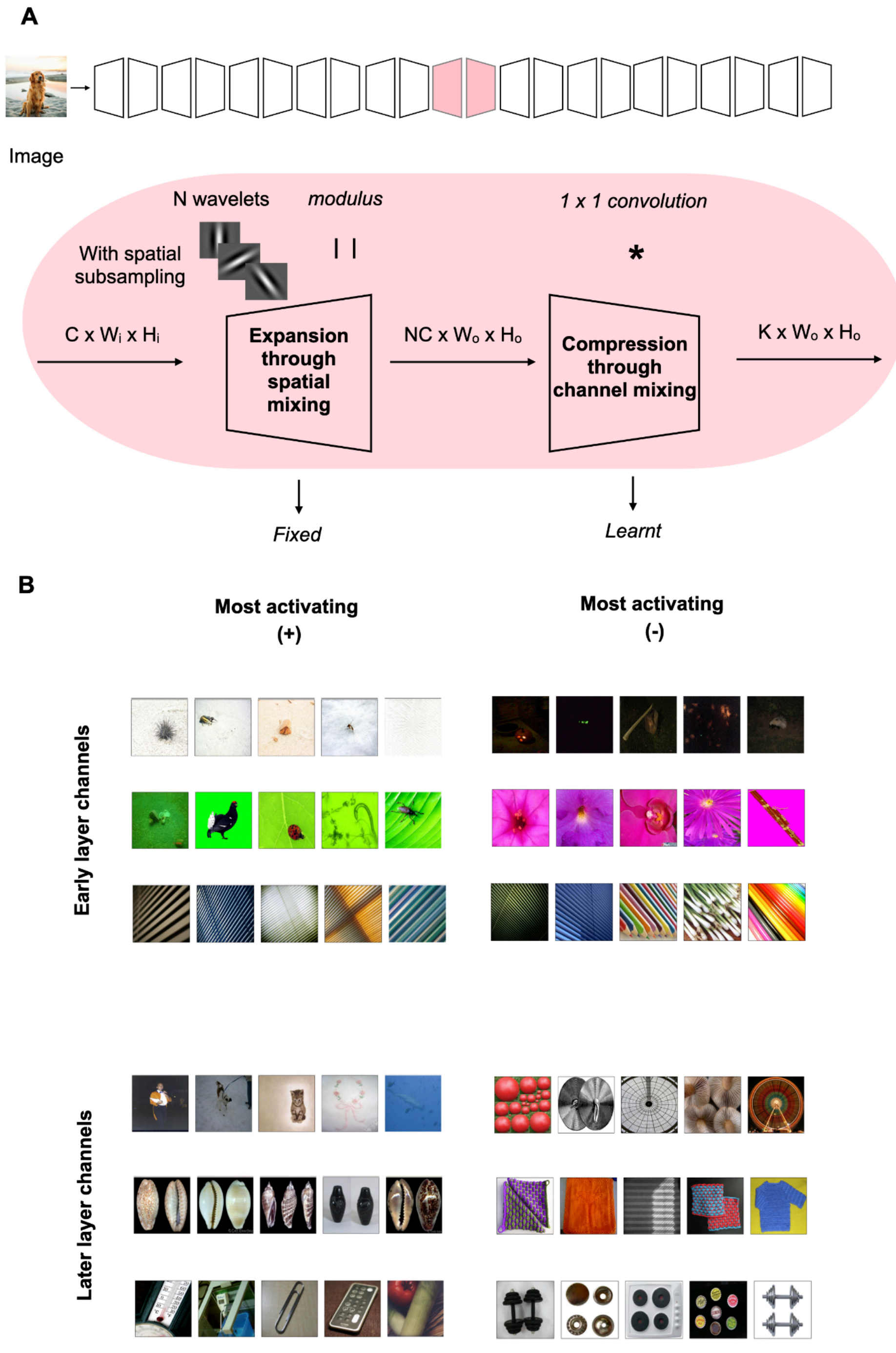

**Figure 1: Hierarchical unsupervised efficient coding yields increasingly complex visual features.** (A) An 11-layer convolutional neural network was developed in which each layer consists of two operations: convolution with a fixed set of N wavelet filters, followed by channel mixing via a learnable 1×1 convolution. The input to each layer has dimensions C × Wi × Hi, where C represents the number of input channels, Wi denotes the input width, and Hi denotes the input height. After expansion through spatial mixing using the N wavelet filters and rectification, the representation grows to dimensions NC × Wo × Ho, where NC is the expanded channel count, and Wo and Ho are the output width and height, respectively. In every other layer,

strided convolution (spatial subsampling) reduces the spatial dimensions by a factor of two. Subsequent 1×1 convolution performs compression through channel mixing, reducing the representation to $K \times Wo \times Ho$, where K is the number of retained principal components. The unsupervised algorithm learns to compress image features into the dominant modes of variation observed in natural images. Specifically, the 1×1 convolution weights are optimized using a sequential, layer-wise principal component analysis on spatially pooled activations from natural images. For each layer, the covariance matrix of these activations is computed, and the top K eigenvectors are extracted to serve as the weights of the 1×1 convolution. This effectively projects the input into a K-dimensional subspace aligned with natural image statistics. In the subsequent layer, the N fixed wavelet filters expand this compressed representation by a factor of N. (B) To visualize feature selectivity, one million images were passed through the model, and the images producing the highest positive and negative activations for each layer were identified. Early layer channels are selective to simple features (such as edges, orientations, and colors), while later layer channels are selective to more complex features, such as objects, textures, and shapes. This progression suggests that hierarchically applying efficient coding principles enables the model to learn increasingly complex representations.

During unsupervised learning, each layer computes the principal components of its input activations using images from the ImageNet training set. Various algorithms have been proposed for efficient coding, including whitening, independent component analysis (ICA), and nonnegative sparse embedding (Beyeler et al., 2019; Karklin & Lewicki, 2003; Olshausen & Field, 1996). However, we chose to use principal component analyses (PCA) for multiple reasons.

First, it has been proposed that while efficient codes in low-level visual processing may be optimized to whiten their inputs, efficient codes in higher-level processing may instead operate in a sampling-limited regime, in which the dominant constraint is the accurate estimation of complex features from finite samples (Hermundstad et al., 2014). It is argued that in this regime, resources should be allocated in proportion to feature variability, because more variable features are more detectable above noise. PCA directly optimizes for maximally varying features. Second, in our model, Morlet wavelet filters already function like the edge detectors that were the focus of previous work (Olshausen & Field, 1996), whereas PCA is used to learn more complex features from the outputs of these edge detectors. Third, although ICA has been argued to be preferable to PCA for modeling V1 (Olshausen & Field, 1996), other work on the principal components (PCs) of Gaussian-windowed natural image patches, a setting similar to ours due to the Morlet wavelets, has shown that groups of PCs together span Gabor-like filters resembling those learned by ICA (Heidemann, 2006). This suggests that after spatial windowing, the second-order statistics of natural images may be sufficient to capture the same structure that ICA learns from higher-order statistics. Fourth, PCA solutions are unique and globally optimal, whereas ICA solutions depend on initialization and algorithm choice. In sum, PCA directly optimizes a theoretically motivated objective for cortical coding, can span the same features learned by ICA after spatial windowing, and is less contingent on implementation details than ICA.

For each layer, activations were collected after wavelet filtering and rectification; their covariance matrix was computed over spatially pooled activations across images. The top K eigenvectors, capturing the dominant modes of natural-image variance, became the weights of that layer's 1×1 convolutions, compressing inputs into a K-dimensional subspace. Subsequently, fixed wavelet filtering and rectification in the next layer then expands these compressed representations, with the expansion factor equal to the number of wavelet filters. Thus, unlike typical deep networks trained end-to-end via backpropagation, our deep efficient coding procedure is a form of layerwise unsupervised learning using only local input statistics, requiring no task labels, no global objective, and no backward pass (Hinton et al., 2006). Full architectural and training details are available in the Methods.

**Deep efficient coding yields behaviorally relevant features**

We first sought to assess the basic feasibility of learning useful high-level representations through efficient coding in a deep feature hierarchy. We began by visualizing the features learned by a model trained only through layerwise efficient coding (Fig. 1B). As expected, the early layers of this model exhibit selectivity for relatively simple features, such as edges, orientations, and colors. However, the later layers exhibit selectivity for much more complex features, including objects, textures, and shapes. This progression from low-level to high-level feature selectivity suggests that applying efficient coding principles hierarchically can enable the emergence of increasingly complex representations, even in the absence of task supervision.

We next wondered whether the high-level representations learned through efficient coding are behaviorally relevant and correspond to features that influence perceptual judgements. To assess this, we conducted behavioral experiments in which participants viewed a target cluster of images and were asked to select which of two option clusters the target most closely resembled. This paradigm is similar to other recent work that examines the interpretability of neural network representations by assessing whether observers or other networks can detect the perceptual similarities among images that strongly activate specific network dimensions (Klindt et al., 2023; Zimmermann et al., 2024). In our experiment, the target and option clusters were constructed based on their activation patterns in individual network channels from the last layer. The selectivity of example channels from this layer are shown in Figure S1. In the task, one option cluster contained images that elicited high activations in a given channel, while the other contained images that elicited low activations. If a channel encodes behaviorally relevant features, participants should reliably select the option cluster that shares the same activation polarity as the target.

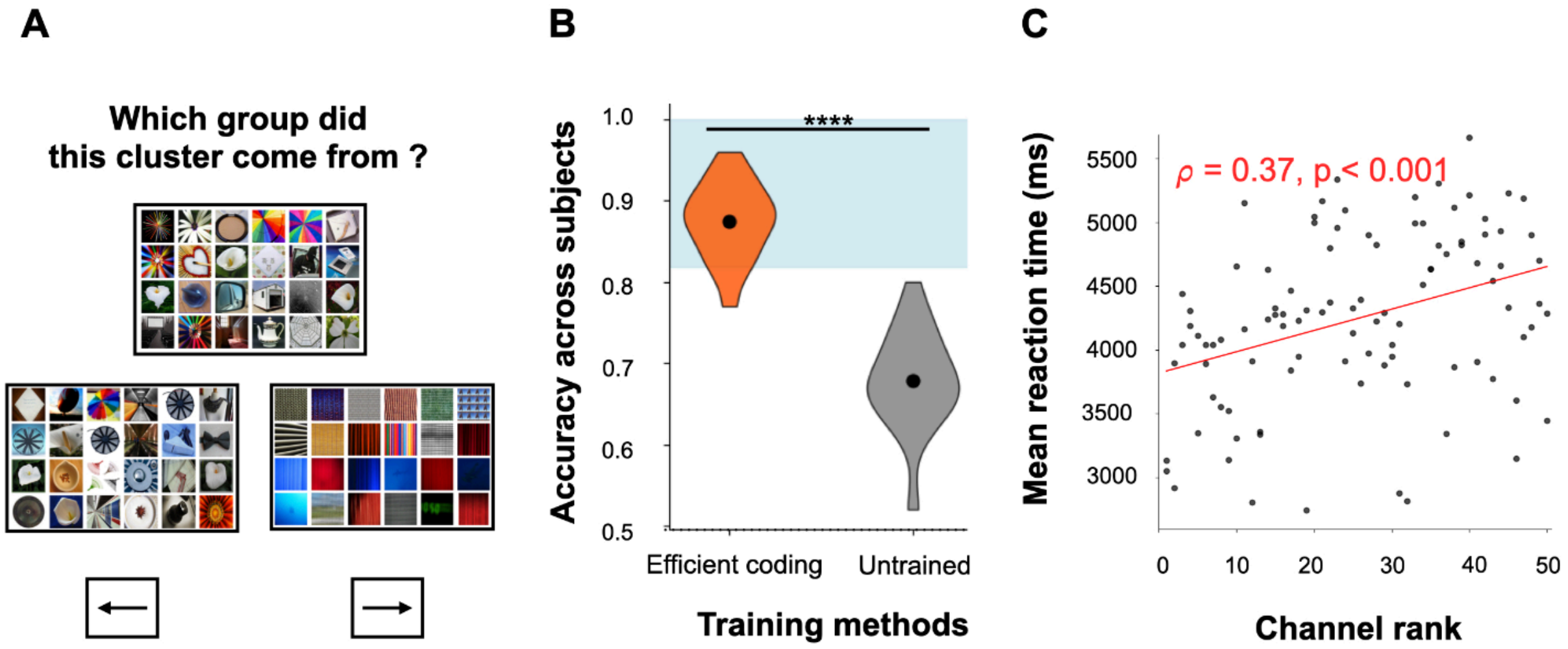

**Figure 2. Deep efficient coding yields behaviorally relevant representations.** (A) Behavioral task for assessing the interpretability of individual network channels. For a given channel in the final layer, the 48 images that produced the highest activations and the 48 that produced the lowest activations were identified. Each set was split into two subsets of 24 images (even vs. odd indices). On each trial, participants saw a target cluster of 24 images (either high-activating or low-activating, from one subset) and two option clusters: one containing high-activating images and the other containing low-activating images. Participants were asked to choose which option cluster best matched the target cluster, providing a behavioral measure of how readily the features represented by each channel could be recognized by human observers. (B) These violin plots show participant accuracy in the interpretability task for a network trained with efficient coding and an untrained network. The blue shaded region indicates the 95% data interval for a fully supervised model trained on one million ImageNet images, providing a point of reference. The unsupervised model exhibits strong accuracy despite training on only 10,000 images, and it significantly outperforms the untrained model. Black dots indicate mean accuracy; violin width reflects the distribution of participant scores. (C) This scatterplot shows the relationship between channel variance rank and mean reaction time for the features of the deep efficient coding model. Each dot represents the mean reaction time for a single channel, averaged across participants. The red line indicates the best-fitting linear regression. Reaction times were significantly correlated with channel rank (Spearman $\varrho = 0.37$, $p < 0.001$), indicating that participants identified features from higher-variance channels more rapidly than those from lower-variance channels. This finding suggests that unsupervised efficient coding organizes representations such that the most statistically prominent dimensions are also the most perceptually salient and recognizable. ****$p<0.0001$

We examined these behavioral effects for an unsupervised deep efficient coding network trained on only 10,000 images from ImageNet. We additionally ran the same behavioral experiment for an untrained version of the same architecture. By matching the architecture across networks, we are able to reveal the specific contribution of efficient coding to the emergence of interpretable features when training with limited data.

The results are shown in Figure 2. First, they demonstrate that both models yield accuracy above chance (50%), indicating that even untrained networks impose some human interpretable structure on image representations. Second, when comparing across networks, we found significantly higher accuracy for the unsupervised model ($M = 0.87$, $SD = 0.33$) relative to an untrained model ($M = 0.68$, $SD = 0.47$; $t = 15.2$, $p < 0.0001$). This demonstrates that deep efficient coding can be used to learn strongly behaviorally relevant representations from just thousands of training images.

To contextualize the performance of the unsupervised network, we also evaluated features from a conventional supervised network trained on the full ImageNet-1K dataset. This model achieves the highest overall accuracy ($M = 0.96$, $SD = 0.2$) and provides a point of reference for the best accuracy achieved when leveraging extensive supervised training using millions of training samples (Fig. 2B). Remarkably, the accuracy for our unsupervised network approaches that of the fully supervised network, despite being trained on only 10,000 images and without any explicit labels. Additionally, as a control, we wanted to determine whether principal components alone were sufficient for behavioral relevance, or whether the hierarchical application of efficient coding was necessary. To test this, we created a model where our PCA learning procedure was applied only to the last layer while the rest of the network remained untrained. This model performed significantly worse than the deep efficient coding model ($M = 0.6$, $SD = 0.5$; $t = 25.2$, $p < 0.0001$), confirming that the hierarchical structure of deep efficient coding is critical for learning perceptually meaningful representations. Together, these results suggest that efficient coding principles are well-suited to support the rapid learning of behaviorally relevant features from limited data.

We next examined reaction time effects. Efficient coding theory predicts that neural resources in higher-level sensory processing should prioritize features capturing the most variance in natural stimuli . If so, we expect higher-variance channels to encode more perceptually salient features. To test this, we examined the relationship between channel variance rank and reaction time for the deep efficient coding model. Consistent with the prediction, reaction times showed a significant positive correlation with channel rank ($\varrho = 0.37$, $p < 0.001$), indicating that participants identified features from higher-variance channels more rapidly than those from lower-variance channels (Fig. 2C). This finding demonstrates that the statistical structure captured by the deep efficient coding network aligns with perceptual salience for human observers, and it is consistent with a version of the efficient-coding theory in which neural representations place more emphasis on the higher variance features of natural stimuli.

### Deep efficient coding enables brain alignment under extreme data limitations

Our behavioral findings suggest that deep efficient coding is effective at learning behaviorally relevant representations from thousands of images. We next sought to determine whether efficient coding principles can also facilitate the emergence of brain-aligned representations under the constraint of limited training data.

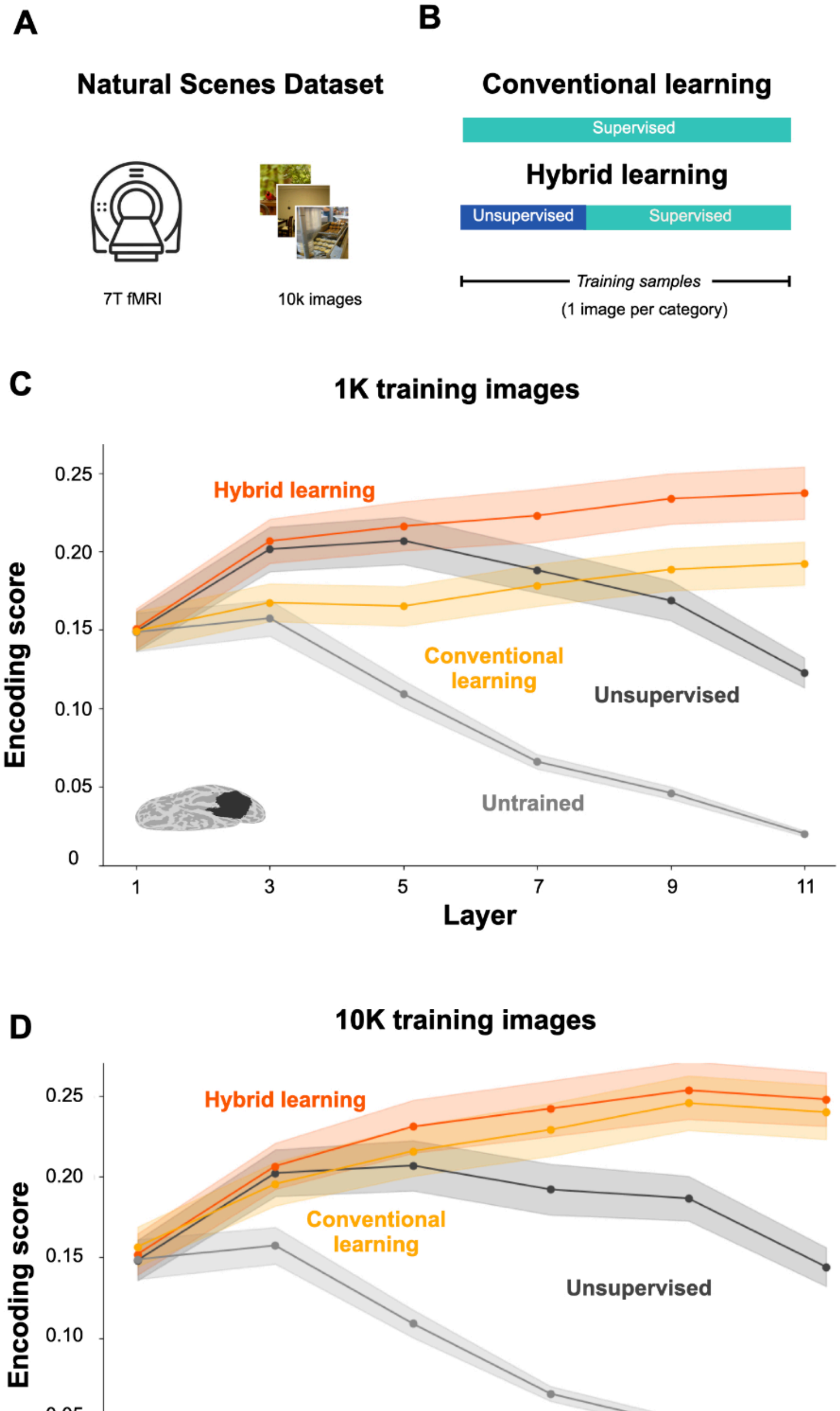

**Figure 3. Hybrid learning that combines efficient coding with supervised fine-tuning improves encoding performance under strong data limitations.** (A) The Natural Scenes Dataset (NSD) comprises high-resolution 7T fMRI recordings from eight subjects viewing ~10,000 natural images spanning diverse scene categories. (B) This schematic illustrates two approaches for training networks on image classification. Conventional learning relies solely on supervised training, while hybrid learning incorporates an initial unsupervised learning phase followed by supervised fine-tuning. (C) This plot shows encoding scores across networks trained on 1,000 images

for hybrid learning (red), conventional learning (yellow), unsupervised learning (black), and an untrained model (gray). Results are shown for every other later across the full network depth. Brain predictivity was evaluated in ventral visual cortex, as indicated by the highlighted region in the brain rendering. With just 1,000 training images, unsupervised learning of a deep efficient code yields a substantial improvement over an untrained version of the same architecture. Furthermore, under these strong data limitations, a hybrid approach that combines efficient coding with supervised fine-tuning yields higher encoding scores than a network trained with conventional supervised learning. Shaded regions indicate the standard error across subjects. (D) This plot follows the same conventions as in panel C but for networks trained on 10,000 images. Again unsupervised learning yields a substantial improvement in encoding performance over an untrained network. However, in contrast to the networks trained with just 1,000 images, training on 10,000 images yields similar performance for the hybrid and conventional models

The representations of the adult visual system are likely shaped by a combination of both unsupervised and supervised learning procedures. Thus, our goal was not to assess whether efficient coding alone could explain representations along the visual hierarchy, but rather to assess whether there may be benefits from incorporating efficient coding principles into supervised networks. To test this possibility, we developed a hybrid-learning approach in which the network first develops a deep efficient code in an unsupervised learning phase followed by a phase of supervised learning on ImageNet classification (Fig. 3B). We first trained this model using an extremely limited training set: just one randomly sampled image from each of the 1,000 ImageNet categories. For comparison, we examined the same architecture trained through conventional supervised learning on ImageNet classification using a matched training set of 1,000 images and a version trained using only unsupervised efficient coding on 1,000 images. We also examined an untrained version of this architecture.

We assessed how well these networks could predict image-evoked responses in the human ventral visual stream using fMRI responses from the Natural Scenes Dataset (NSD) (Fig. 3A) (Allen et al., 2022). This dataset is the largest existing human fMRI dataset on natural scene perception, containing responses to ~73,000 images across 8 participants. Using regularized linear regression, we fit encoding models that mapped layer-wise representations from each neural network to the voxelwise fMRI responses. These encoding models were estimated on a set of ~9,000 training images, and their performance was evaluated on a held-out set of ~1,000 test images. Encoding performance was measured as the correlation between predicted and actual cortical responses in the test set.

We examined the average encoding performance in a region of interest (ROI) in the high-level ventral visual stream (Fig. 3C-D). First, the findings show that unsupervised learning alone yields a major improvement relative to an untrained version of the same architecture. Second, they show that a hybrid approach that combines unsupervised and supervised learning yields a further improvement over unsupervised learning alone, specifically in later network layers (i.e., layer 6 and above). Third, the advantage of hybrid learning over conventional supervised learning depends on the amount of available labeled training data (Fig. 3C-D). When the network training data are restricted to just 1,000 images, hybrid learning substantially outperforms conventional supervised learning, with particularly pronounced differences in later layers. However, when the training data are increased to 10,000 images, hybrid and conventional learning achieve comparable encoding scores. Together, these results demonstrate that efficient coding alone yields a strong improvement in encoding scores relative to an untrained baseline and that a hybrid approach that combines efficient coding with supervised fine-tuning is beneficial when labeled data are scarce.

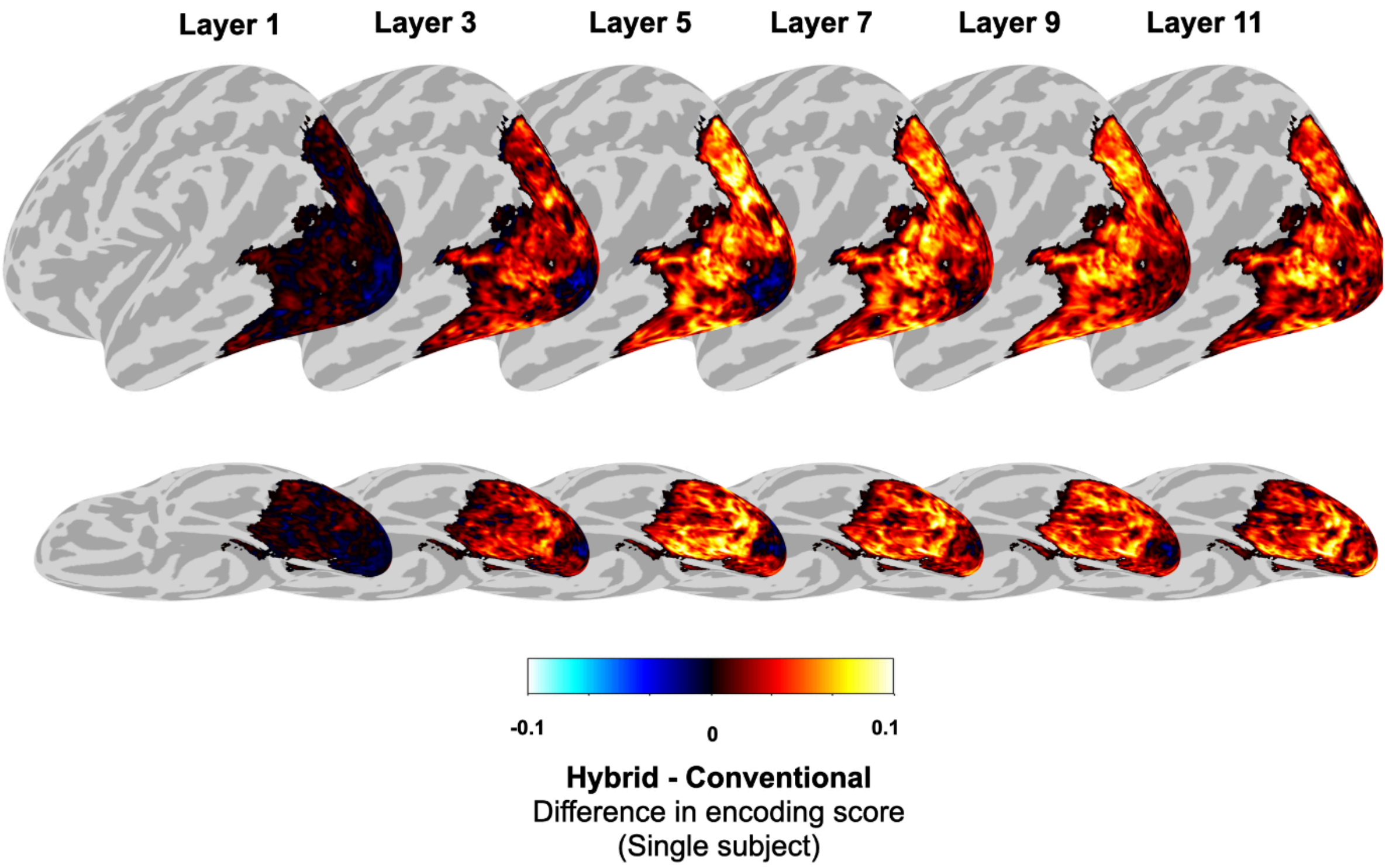

**Figure 4: Hybrid learning improves encoding performance across visual cortex when training data are limited.** Cortical surface maps show the difference in encoding scores (Hybrid − Conventional) for a representative subject across network layers, for networks trained on 1,000 images. Warm colors indicate voxels where hybrid learning outperforms conventional training; cool colors indicate the reverse. Top row: lateral view; bottom row: ventral view. The widespread positive values across visual cortex demonstrate consistent improvements, particularly in mid-to-late layers of the network.

We next examined voxelwise effects to determine whether the performance benefits of the hybrid model in low-data settings were concentrated in specific regions or widely distributed. We plotted the effects in a large region of visually responsive voxels using the “nsdgeneral” ROI provided by the original authors of the NSD study. Figure 4 shows the difference in voxelwise encoding scores for hybrid learning relative to conventional supervised learning when training on 1,000 images. These effects are shown for layers sampled along the entire hierarchy of the network. Figure S4 shows the voxelwise encoding performance for both the models separately. Consistent with the ROI results in Figure 3C, these plots show that hybrid learning yields a substantial improvement in encoding performance relative to conventional supervised learning, but they further show that these improvements are found throughout the visual cortex and are not driven by a small subset of visual regions.

Together these findings show that deep networks trained using a hybrid approach that combines efficient coding with task supervision learn brain-aligned representations from a strikingly small sample of images, as little as one image per ImageNet category.

**Deep efficient coding facilitates downstream object classification**

Our results thus far demonstrate that initializing a network with a hierarchical efficient code yields behaviorally representations and improves brain alignment when training data are scarce. We next wondered if the representations learned through hierarchical efficient coding might also facilitate downstream learning on an image classification task. To test this, we trained networks to perform object classification using the miniImageNet dataset (Vinyals et al., 2016). We compared the training dynamics of conventional supervised learning with a hybrid approach that combines deep efficient coding with subsequent supervised learning.

In the hybrid model, the network first develops a deep efficient code through unsupervised learning using 10,000 images randomly sampled from ImageNet. The network is then trained through supervised learning on the object classification task. In conventional learning, the network is trained on object classification from a random initial state. As shown in Figure 5A, we find that hybrid learning improves the training dynamics. Networks that are initialized with a deep efficient code require fewer supervised training samples to reach the same level of performance as a conventional supervised network. Note that the difference between the hybrid and conventional networks cannot be simply explained by the total number of training images after accounting for the unsupervised phase of the hybrid model. This is because the number of training samples plotted on the x-axis in Figure 5A already takes the unsupervised phase into account. We confirmed that these results are not due to atypical learning dynamics in the scattering network architecture compared with more conventional architectures: as shown in Figure S2, the supervised learning dynamics are similar to those for ResNet50. Furthermore, a control experiment showed that the improvements for the hybrid model are specifically driven by the features learned in the unsupervised phase and not by a more general property of the weight distributions. Specifically, we found that if we permute the 1x1 convolutional filter weights after unsupervised learning, the training dynamics resemble those for conventional learning (Fig. S3).

The analyses thus far show that deep efficient coding provides an advantageous starting point for learning to classify objects. We next performed an even stronger test of how well deep efficient coding can support task performance: we froze the early to intermediate layers of the hybrid network after unsupervised learning and only allowed the later layers to be updated through task supervision on miniImageNet. The goal was to determine if low- to mid-level features learned through an efficient-coding objective alone might be sufficient to support downstream object classification. For comparison, we performed the same layer-freezing experiments on a conventional supervised network, with early layers frozen in their initial random state and with task-supervised weight updates only applied to the remaining layers. We performed these analyses with freezing up to layer 5, since our previous encoding-model results suggest that for the first five layers, efficient coding alone may be as effective as hybrid learning (Fig. 3C).

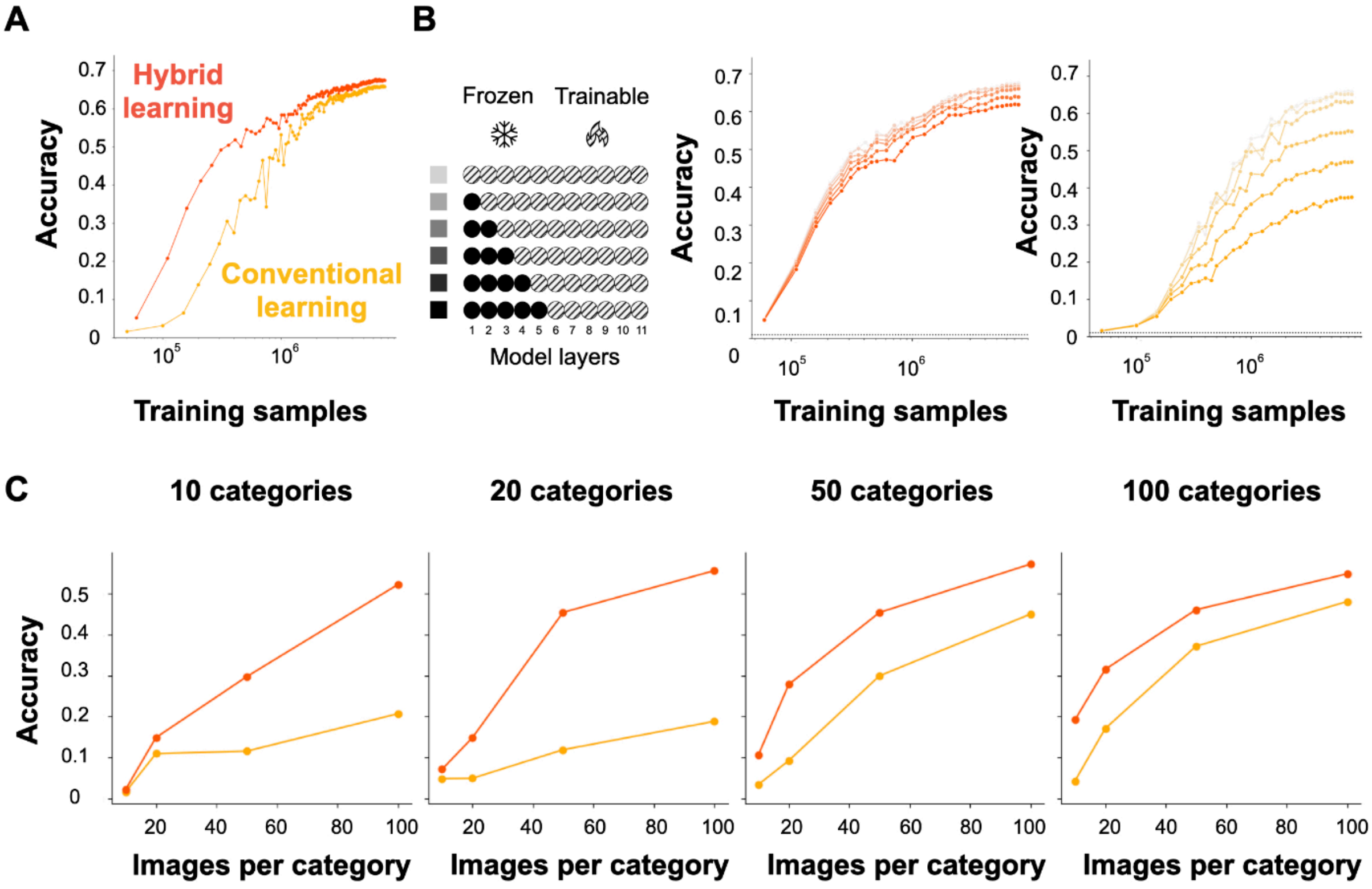

**Figure 5. Unsupervised efficient coding enables data-efficient learning.** (A) Classification accuracy on miniImageNet as a function of the number of training samples for models trained with conventional supervised learning (yellow) or a hybrid approach that incorporates efficient coding principles in an unsupervised pre-training phase (red). Hybrid models learn faster and achieve higher accuracy with fewer training samples. (B) Transferability of early network layers assessed by freezing a varying number of early convolutional layers while training the remaining layers on object classification. The schematic (left) illustrates the freezing protocol across the 11-layer network. Line opacity indicates the number of frozen layers, with the least opaque lines representing end-to-end trained models (no layers frozen) and the most opaque lines representing models with half of the layers frozen. Classification accuracy is plotted as a function of training samples for hybrid (left, red) and conventional (right, yellow) learning. Models pre-trained with unsupervised learning maintain high classification accuracy even when up to half of the layers are frozen, whereas conventionally trained models show pronounced performance declines when layers are frozen. (C) Classification accuracy for hybrid (red) and conventional (yellow) learning when training on varied numbers of categories and samples per category. The advantage of unsupervised pre-training is most pronounced in data-limited settings.

As shown in Figure 5B, the performance of the hybrid model is remarkably robust to layer freezing. Even when nearly half of the layers are frozen in their unsupervised state, the effect on classification performance is relatively minor. In contrast, applying the same freezing protocol to randomly initialized networks has a much larger effect, with a major drop in performance when the first five layers are held fixed. This demonstrates that the unsupervised learning phase is crucial for obtaining strong classification performance when early to intermediate layers are frozen. As in the preceding analyses, these effects cannot be simply explained by the total number of training images, because the number of training samples plotted on the x-axis in Figure 5B already takes the unsupervised phase into account.

**Deep efficient coding facilitates learning in low-data regimes**

Our findings suggest that a major strength of deep efficient coding is the ability to learn useful features in a data-efficient manner. We wondered if this strength could be leveraged to support category learning from little data—a skill that humans are adept at. We systematically tested how unsupervised pre-training affects category learning across a range of data-limited conditions. We compared two training approaches: conventional learning, in which networks were trained from random initialization using only supervised weight updates, and hybrid learning, in which networks first underwent unsupervised pre-training before supervised fine-tuning. Critically, both approaches used identical architectures, training images, and supervised learning procedures, allowing us to isolate the effect of unsupervised pre-training. Both approaches used identical training images. Models were trained using a varied number of categories (10, 20, 50, or 100) and a varied number of training images per category (10, 20, 50, or 100).

The results are shown in Figure 5C. Across all data-limited settings that we examined, networks trained using hybrid learning consistently outperform those trained with conventional supervised learning, but the difference between the hybrid and conventional approaches diminishes as the number of training categories becomes large. In contrast, when models are trained on a more limited number of categories, the advantage of unsupervised pre-training is pronounced, allowing the network to rapidly learn category distinctions and yielding substantially higher performance than a fully supervised approach. This suggests that unsupervised pre-training gives a boost to category learning in settings where the training data and the number of available category labels are limited.

We next performed exploratory visualization analyses to examine how unsupervised pre-training boosts category learning. Specifically, we used uniform manifold approximation and projection (UMAP) to generate two-dimensional embeddings of the layer 11 representations from networks before and after supervision (Fig. 6). We examined the conventional and hybrid networks from Figure 5C trained on ten categories with 100 images per category. For the conventional network, the representations before supervision are in a random initialized state and show little category clustering. In contrast, the hybrid network, which has undergone unsupervised pre-training, already exhibits some clustering of perceptually similar images before any supervised training. After supervised training, both the conventional and hybrid networks show category clustering, but the initial perceptual organization of the hybrid model results in superior final classification performance, particularly in low-data settings.

Together, these results demonstrate that unsupervised efficient coding provides a powerful foundation for rapid category learning. By organizing hierarchical representations along high-variance axes of natural images before supervised training, efficient coding enables networks to achieve higher classification accuracy with fewer training samples, an advantage that is most pronounced when training data are limited.

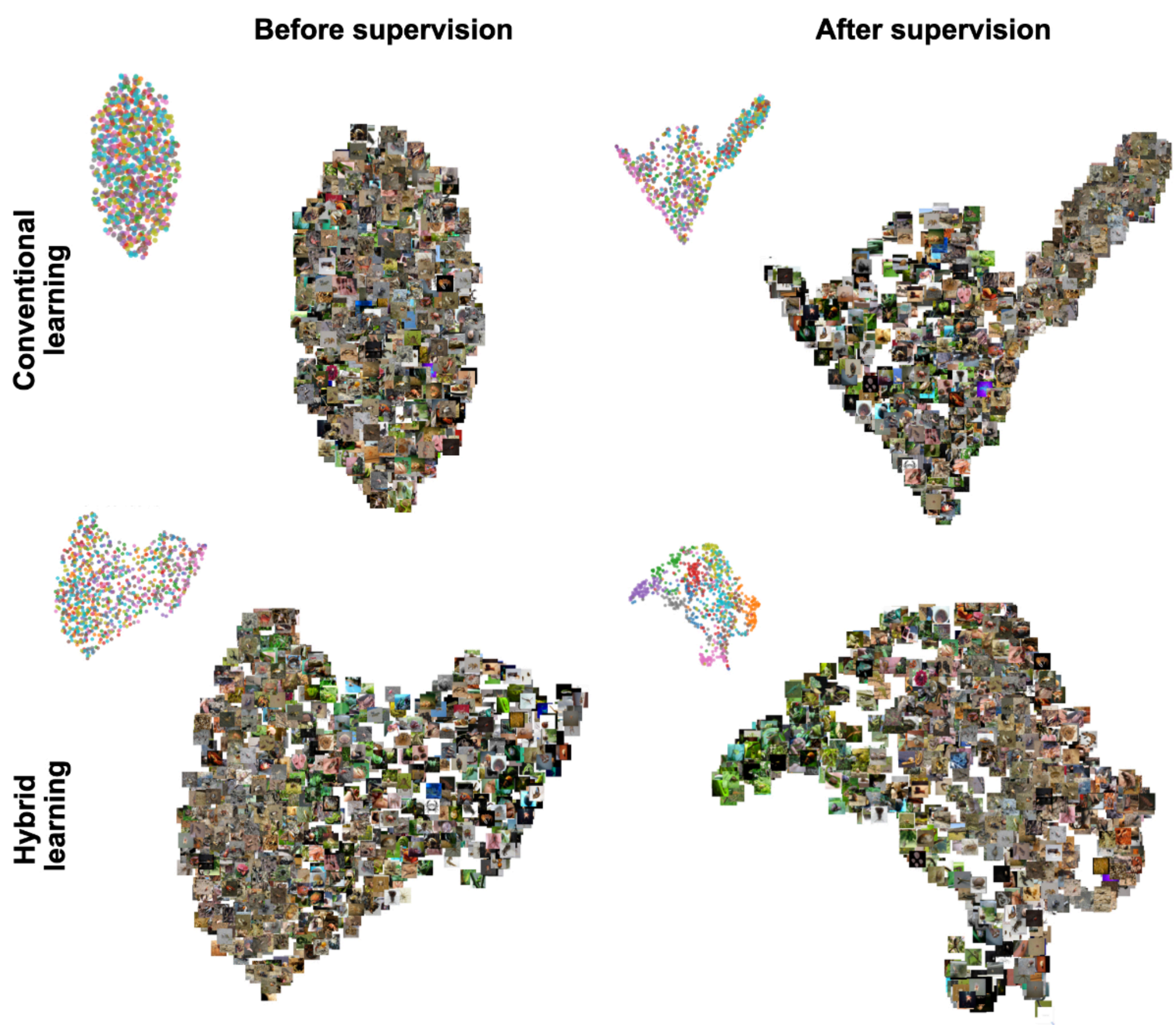

**Figure 6: UMAP visualizations of training set activations before and after supervision for models trained with conventional and hybrid learning.** At initialization, the conventional network contains largely unstructured image representations (top left), with category clustering emerging only after extensive supervised training (top right). In contrast, the unsupervised phase of the hybrid network yields perceptual organization and discernible

clusters even before supervised training (bottom left), suggesting that unsupervised pre-training facilitates the extraction of useful, task-relevant features.

## DISCUSSION

We developed an unsupervised efficient coding procedure that tunes a deep hierarchy of features to the dominant modes of variation in natural images. We found that the features learned from deep efficient coding are behaviorally relevant, predict high-level visual cortex representations, and facilitate downstream classification performance. We further show that a hybrid approach combining efficient coding with supervised fine-tuning yields brain-aligned features from extremely limited data, as few as one image per ImageNet category. Together, these findings suggest that efficient coding principles may shape representations across the entire visual hierarchy and could be crucial for understanding the remarkable data efficiency of biological vision.

Our work suggests a new theoretical perspective on brain-aligned representations in deep vision models. Early goal-driven networks trained on supervised image classification successfully predict ventral stream responses (Khaligh-Razavi & Kriegeskorte, 2014; Yamins et al., 2014), and more recent self-supervised models achieve similar cortical alignment (Konkle & Alvarez, 2022; Zhuang et al., 2021). However, these approaches all rely on task objectives and end-to-end backpropagation through deep architectures. Our framework departs from this paradigm in three important ways. First, the unsupervised efficient coding stage uses no task and no backpropagation. Learning arises purely from the statistical structure of natural images through local learning mechanisms. Second, this unsupervised learning alone produces genuinely useful features: when early and intermediate layers are frozen after efficient coding and only later layers receive task supervision, classification performance remains strong, whereas freezing the same layers in randomly initialized networks causes dramatic performance drops. Third, our hybrid approach combining efficient coding with supervised learning achieves strong brain predictivity with a remarkably small training set. Together, these results suggest an alternative hypothesis for the optimization constraints shaping ventral visual cortex. In contrast to the prevailing approach of end-to-end supervised or self-supervised learning, we suggest that across the ventral visual hierarchy, representations are shaped by both bottom-up efficient coding and top-down goal-driven learning.

Many early studies have implemented efficient coding principles in relatively shallow models (Karklin & Simoncelli, 2011; Olshausen & Field, 1996, 1997, 2004). More recent efforts have begun extending these principles along a deep hierarchy of visual processing (Yerxa et al., 2020, 2023) and our work contributes to this emerging direction. By stacking multiple layers of PCA-based compression, our model learns progressively more complex representations, enabling it to predict neural responses in higher-level visual areas. Additionally, traditional efficient coding work has focused on developing optimal sensory representations as an end in itself. Here, we demonstrate that hierarchical efficient coding can serve as a powerful initialization for supervised learning, accelerating convergence and improving data efficiency. Importantly, however, unsupervised efficient coding alone is not sufficient; supervision remains necessary to achieve strong classification performance and neural predictivity. This suggests that efficient coding and task-driven learning are complementary: unsupervised exposure to natural image statistics builds a foundation of reusable features, while supervised training refines these representations for behaviorally relevant goals.

Our approach also connects to classic work on layer-wise pre-training in deep networks. Early greedy layer-wise methods showed that training each layer locally (without end-to-end supervision) yields reusable hierarchical features and improves downstream optimization and generalization (Bengio et al., 2006; Erhan et al.). Large-scale experiments further demonstrated that unlabeled data alone can give rise to high-level feature detectors useful for recognition (Le et al., 2012). Our layer-by-layer PCA procedure can be viewed as a particularly simple instance of such greedy training, replacing reconstruction with a local variance-maximization rule. Moreover, the emergence of higher-order selectivity we observe aligns with theory on deep linear networks, which predicts that gradient descent learns input–output modes in order of their singular values (Saxe et al., 2014), producing efficient codes with greater sensitivity to high variance features (Benjamin et al., 2022). Together, these results suggest that hierarchical, locally trained systems that prioritize high-variance directions can develop transferable, brain-like features and achieve strong data efficiency without global backpropagation.

There are a few important caveats to our implementation. First, although we organize the network into expansion–contraction stages, these are introduced purely as computational scaffolding to realize our local learning rule, not as literal models of cortical anatomy. Recent work shows that diverse deep-network architectures converge on a common set of feature dimensions when trained on natural images, suggesting that the learned features are more important than the details of the network architecture (Chen & Bonner, 2024). Second, our decision to fix a bank of spatial filters and learn only the channel-mixing weights departs from conventional convolutional networks, where both spatial and channel weights are trained. This design choice was made so that the model could discover an optimal basis for combining feature channels, while relying on a known efficient spatial basis, thereby simplifying learning. Lastly, we chose PCA as our dimensionality reduction method because it directly optimizes for high-variance features and yields unique, globally optimal solutions. Future work could explore whether alternative methods such as ICA or nonnegative sparse embedding yield stronger alignment or more interpretable features.

Our model makes several testable predictions. When subjects are exposed to new sensory statistics without any task-based or supervised feedback, our theory predicts that representations across the visual hierarchy may automatically adapt to the highest-variance features in the new stimuli. Given the data efficiency observed here, this unsupervised adaptation might occur rapidly, potentially on the timescale of minutes to hours. These predictions could be tested with psychophysical experiments where subjects are passively exposed to structured visual input and later assessed on discrimination tasks targeting features that varied to different degrees during exposure.

Beyond neuroscience, our findings may have practical applications for computer vision. Our results show that if a network's early to intermediate layers are pre-aligned to natural image statistics, fewer labeled examples are needed to train the network on a downstream supervised task. This suggests that incorporating efficient coding principles into deep learning algorithms may have potential value in settings where labeled data are scarce or expensive to obtain.

To summarize, our work demonstrates that a simple, local, unsupervised algorithm can rapidly carve out hierarchical representations that are behaviourally relevant, neurally aligned, and useful for downstream tasks. These results are consistent with evidence of “unsupervised pre-training” in visual cortex (Zhong et al., 2025), where brain circuits appear to fine-tune themselves through mere exposure alone, suggesting that our fully unsupervised approach may reflect a core biological learning principle that requires neither labels nor rewards.

## METHODS

### Behavioural Experiment

We performed behavioral experiments to assess whether human observers could readily detect the features represented by the 1x1 channels in layer 11. This experiment was performed on several models with the same scattering-network architecture: an unsupervised efficient coding model trained on 10,000 images, an untrained model, a supervised model trained on ImageNet classification using one million images, and a model with unsupervised learning applied only to the last layer (instead of all layers).

To visualize the representations of the 1x1 channels, we processed one million ImageNet images through each network and selected the 48 highest and lowest activating images for each channel. From these, we created two disjoint sets of 24 images for each pole (i.e., high vs. low) by splitting into even- and odd-ranked images. For the experiment, participants saw three clusters of 24 images presented on the screen and had to indicate by button press which of two "option" clusters at the bottom best matched the "target" cluster at the top. On each trial, the target cluster contained images from one pole of a channel, and the option clusters contained either other images from the same pole or images from the opposite pole. We hypothesized that if a channel is selective for behaviorally relevant features, participants would consistently select the option with the same activation polarity (top or bottom) as the target cluster.

We conducted these behavioral experiments on the Prolific online platform. Each experiment used feature representations derived from a different network. All networks share the same architecture but vary in training: unsupervised with 10,000 images (number of subjects = 45), untrained (number of subjects = 46), supervised with one million images (number of subjects = 48), and unsupervised in the last layer only (number of subjects = 44). For each model, we examined the feature selectivity of the 50 channels with the highest variance. Each channel contributed two trials (one with top-activating images as the target cluster and one with bottom-activating images), yielding a total of 100 trials per experiment. Additionally, we included 4 catch trials in each experiment, in which one of the option clusters was identical to the target cluster. Subjects who did not achieve 100% accuracy on the catch trials were excluded from subsequent analysis. After this exclusion, the final number of subjects for each experiment was as follows: 35 for unsupervised, 35 for untrained, 38 for supervised, and 33 for unsupervised in the last layer only. For reaction time analyses, we excluded trials with response times less than 200 ms or greater than 10 seconds (with a median of 7 trials excluded per subject).

### Model architecture

Our network had 11 layers with channel dimensions matching Guth et al. (2024) except for the first layer. Our channel dimensions were [27, 64, 64, 128, 256, 512, 512, 512, 512, 512, 256]. Guth et al., (2024) used 32 channels in the first layer, whereas we used 27 because the input RGB image (3 channels) convolved with 9 wavelet filters produces 27 channels, and our PCA approach cannot produce an overcomplete basis. In each layer, the input first undergoes convolution with a fixed set of nine wavelet filters—eight Morlet band-pass (ψ) filters at a single scale and at eight orientations, along with one low-pass (φ) filter. The φ filter outputs capture coarse structure and pass linearly, whereas ψ filter outputs detect oriented bandpass features and are rectified via the complex modulus. In every other layer, strided convolution (spatial subsampling) reduces the spatial dimensions by a factor of two, which means that deeper layers detect features over an effectively larger receptive field. All filter responses go through batch-normalization, which subtracts the running mean and divides by the running standard deviation to produce zero-mean, unit-variance outputs. The learnable 1×1 convolutions mix these filter responses at each spatial location. We then apply spatial $L_2$-normalization: at every spatial position we treat the vector of mixed features as an $n$-dimensional arrow and rescale it to unit length, ensuring that no single location can dominate purely by having larger magnitude. By repeating these basic operations across 11 layers, the network builds a deep hierarchy of feature representations.

**Unsupervised learning protocol**

Through unsupervised learning, we optimized only the 1×1 channel-mixing weights using sequential, layer-wise PCA on activations extracted from images sampled from the ImageNet training set (Krizhevsky et al., 2012). For each layer, activations were collected after wavelet filtering, rectification, batch normalization, and global average pooling; their covariance matrix was then computed over images. The top K eigenvectors (capturing the dominant modes of natural-image variance) became the weights of the layer's 1×1 convolution, compressing inputs into a K-dimensional subspace. The number of feature maps then expands again in the next layer by a factor of nine, due to the nine fixed wavelet filters.

For networks trained using only unsupervised learning, we performed training using 10,000 images randomly selected from ImageNet. We selected this training set size based on preliminary analyses of encoding model performance, which showed that networks trained on 10,000 images achieved comparable performance to those trained on substantially larger sets of 50,000 and 100,000 images, suggesting that the PCA-based optimization converges to stable feature representations with relatively small data requirements. This data efficiency may be due to the fact that the scattering network has a much smaller number of learnable parameters than conventional deep neural networks. Additionally, we confirmed that models trained on different random samples of 10,000 images exhibit similar results (Fig. S5), indicating that our findings are not contingent on the specific choice of training images.

In subsequent experiments involving hybrid models that incorporate both unsupervised and supervised learning, we trained models using either 1,000 or 10,000 images. For these experiments, both the unsupervised and supervised learning phases were performed using the same image set, which was created by randomly selecting either 1 or 10 images per the 1,000 ImageNet categories.

**Supervised learning protocol**

We trained our scattering network architecture on image classification using two learning paradigms: conventional learning and hybrid learning. In both cases, the architecture consisted of a feature extraction backbone followed by a classification head, and the wavelet filters used for spatial mixing remained fixed throughout training. The two paradigms differed only in how the channel mixing weights were initialized before the supervised training began. In conventional supervised learning, these weights were randomly initialized. In hybrid learning, the channel mixing weights were initialized through unsupervised efficient coding before being further updated during supervised training. Input images were resized and cropped to 224×224 pixels. During training, we applied random resized cropping and random horizontal flipping for data augmentation, while validation images underwent standard resizing followed by center cropping. Models were trained for 150 epochs using the AdamW optimizer. We employed a cosine learning rate schedule with an initial learning rate of $10^{-3}$, preceded by a warm-up period of 10,000 steps. Weight decay was set to $10^{-4}$ to provide regularization. Cross-entropy loss was used as the training objective, and classification performance was evaluated using accuracy on the validation set. Training progress was monitored by logging model evaluation at the end of each epoch. Model checkpoints were saved at predetermined epochs to enable analysis of learning dynamics throughout training.

**Classification analysis**

To understand the differences in features learned through hybrid learning and conventional learning, we used the miniImageNet dataset because it contains sufficient class diversity and image complexity to reveal meaningful differences between learning paradigms without requiring the computational costs of full-scale ImageNet training. To compare the learning dynamics of hybrid and conventional learning, we trained models using each method and computed classification accuracy at every epoch. We plotted accuracy as a function of the cumulative number of images the model had been exposed to during training. For the hybrid approach, this count included both the images used during the unsupervised efficient coding phase and those used during subsequent supervised learning, allowing for a fair comparison of sample efficiency between the two paradigms.

**Layer freezing experiments**

To investigate the contribution of unsupervised efficient coding initialization to learning dynamics, we conducted layer freezing experiments on miniImageNet classification in which varying numbers of early backbone layers were frozen during supervised training. We systematically froze an increasing number of layers starting from the first layer up to the fifth. We performed these analyses with freezing up to layer 5 because our encoding-model results suggest that for the first five layers, efficient coding alone may be as effective as hybrid learning (Fig. 3C). We compared two initialization conditions: (1) channel mixing weights initialized through efficient coding applied to natural image statistics, and (2) randomly initialized channel mixing weights. In all conditions, the frozen layers retained these initial weights throughout training while the remaining layers were updated through backpropagation.

All other aspects of the training protocol remained identical to the standard training procedure. Models were trained for 150 epochs with the AdamW optimizer. We employed a cosine learning rate schedule with an initial learning rate of $10^{-3}$, preceded by a warm-up period of 10,000 steps. Weight decay was set to $10^{-4}$ to provide regularization. These experiments allowed us to assess how much task-relevant information was captured at different depths of the network by the unsupervised initialization alone, independent of subsequent supervised fine-tuning.

**Classification with limited data**

To compare the effectiveness of hybrid learning and conventional learning in low-data settings, we evaluated performance on classification tasks with different numbers of randomly sampled ImageNet categories (10, 20, 50, and 100), and different numbers of images per category (10, 20, 50, and 100). For both approaches, models were trained using the same dataset. Importantly, all models in this section had exactly the same architecture: 27 channels in the first layer and 64 channels in all subsequent layers. Note that we needed to reduce the number of channels in deeper layers relative to the architecture used in our other experiments. This is because some of the datasets examined here had very small numbers of training images (as few as 100), which restricts the numbers of PCs we can learn. In the hybrid learning approach, unsupervised efficient coding was carried out using the exact same images as those used in the supervised phase. The test set for all conditions consisted of 100 images per category.

## fMRI data

For our analysis, we used human fMRI data from the publicly available Natural Scenes Dataset (Allen et al., 2022), which includes neural responses to 73,000 natural images sourced from the Microsoft COCO dataset (Lin et al., 2014). The data were acquired at ultra-high field strength (7T) with high spatial resolution (1.8 $mm^3$ voxels, 1.6-second TR). Each image was displayed for three seconds while participants maintained visual fixation. We utilized single-trial beta estimates provided by NSD, which were preprocessed in 1.8-mm volume space and denoised using the GLMdenoise method, specifically the “betas_fithrf_GLMdenoise_RR” data (Kay et al., 2013). These beta values were standardized (converted to z-scores) within each scanning session, and we averaged responses across repeated presentations of each stimulus. Each participant viewed up to 10,000 images (though the exact number of stimuli seen by each subject varied from 9,209 to 10,000). These 10,000 images were composed of 1,000 images that were common across all subjects and 9,000 that were unique to each subject. For our ROI analyses, we selected visually driven voxels in the ventral stream by taking the intersection of the “ventral stream” and “nsdgeneral” parcels provided by the authors of the NSD data (nsdgeneral identifies voxels with robust image-evoked responses).

## Encoding models

We applied linear regression with L2 regularization to predict voxel responses based on model activations. The dataset was divided into training and test images. In the fMRI data, each participant's training set consisted of images they saw uniquely (up to 9,000), whereas the test set included the common images that all participants viewed (up to 1,000). To determine the most effective mapping for each model, we used leave-one-out cross-validation on the training images to find the optimal L2 penalty for each participant. We evaluated 10 L2 penalty values that were logarithmically spaced using base 10, ranging from 1 to $10^{10}$, and we selected the value that delivered the best average performance across all voxels. With the optimal penalty chosen, we then estimated the regression coefficients using the entire training set. These coefficients were applied to the model activations of the held-out test images to generate predicted neural responses. Finally, we measured the accuracy of these predictions by calculating the Pearson correlation coefficient between the predicted responses and the actual responses for each voxel.

## SUPPLEMENTARY FIGURES

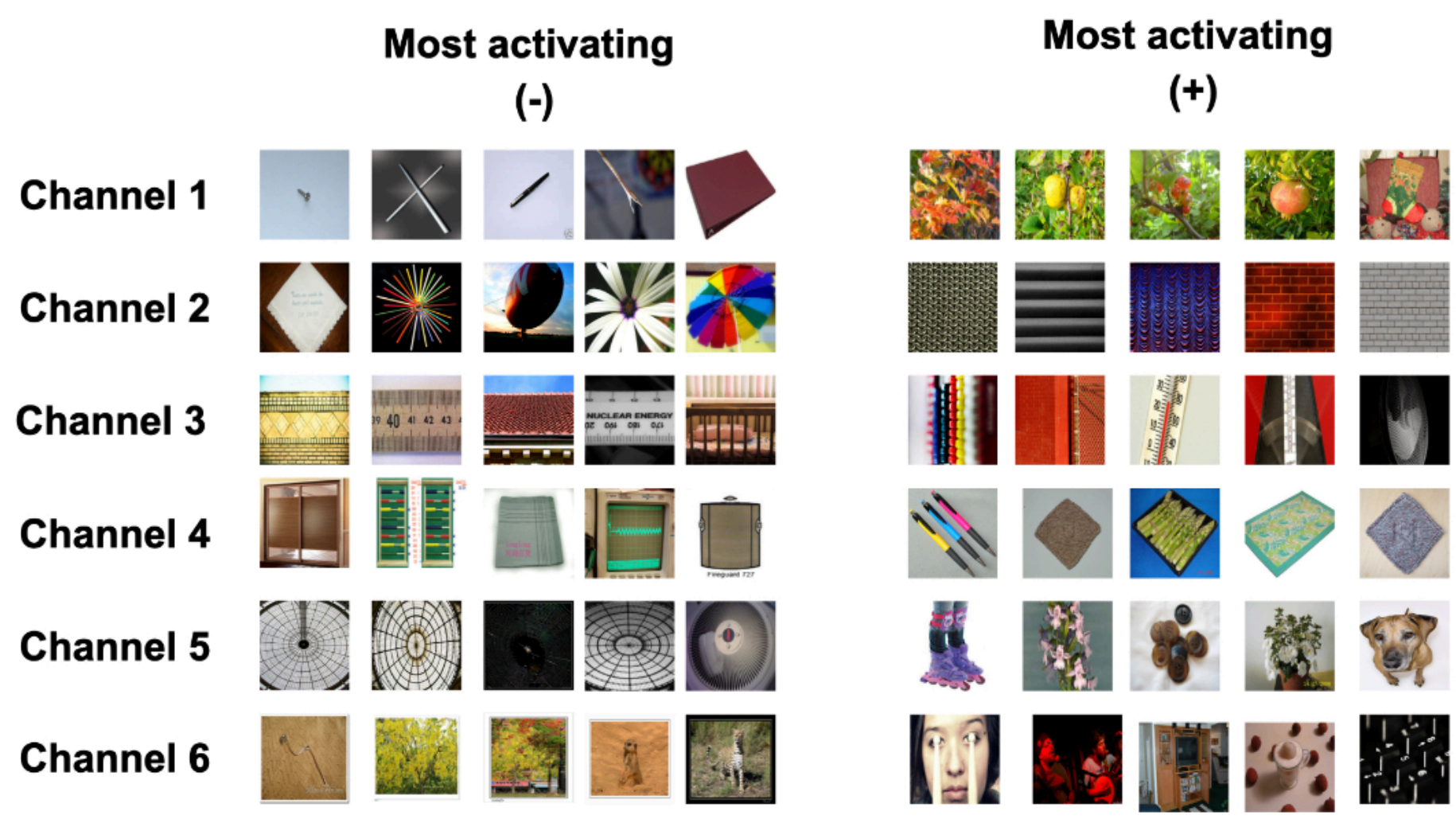

**Figure S1: Image clusters that maximally activate individual network channels in the unsupervised efficient coding model.** For each example channel, images are shown that elicit the strongest negative (*left*) and positive (*right*) activation values. These clusters were constructed by passing 1 million ImageNet images through the network and selecting images with top activation polarity in each channel. Later they were used in behavioral experiments to assess the interpretability of learned representations. Participants viewed a target cluster (either negative or positive activation) and selected which of two option clusters most closely resembled the target, revealing which channel features are perceptually meaningful to human observers

.

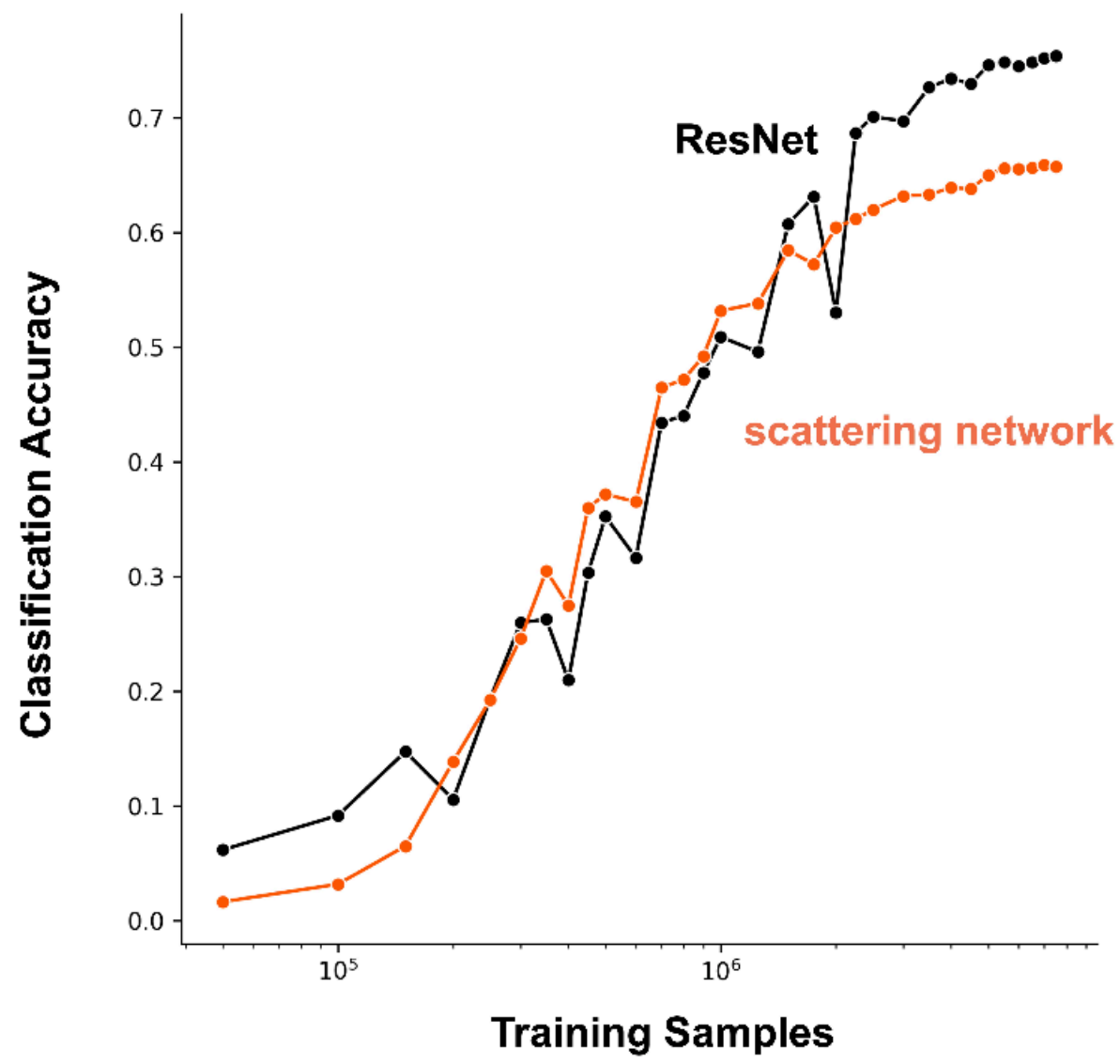

**Figure S2: Scattering networks have a similar training trajectory as ResNet50.** Classification accuracy (y-axis) was tracked as a function of the number of training samples (x-axis) for two very different architectures: a standard ResNet50 and a scattering network. Despite their architectural differences, both networks learn at the same rate, indicating that changing from a deep convolutional backbone to a scattering-based backbone does not strongly alter the overall training dynamics.

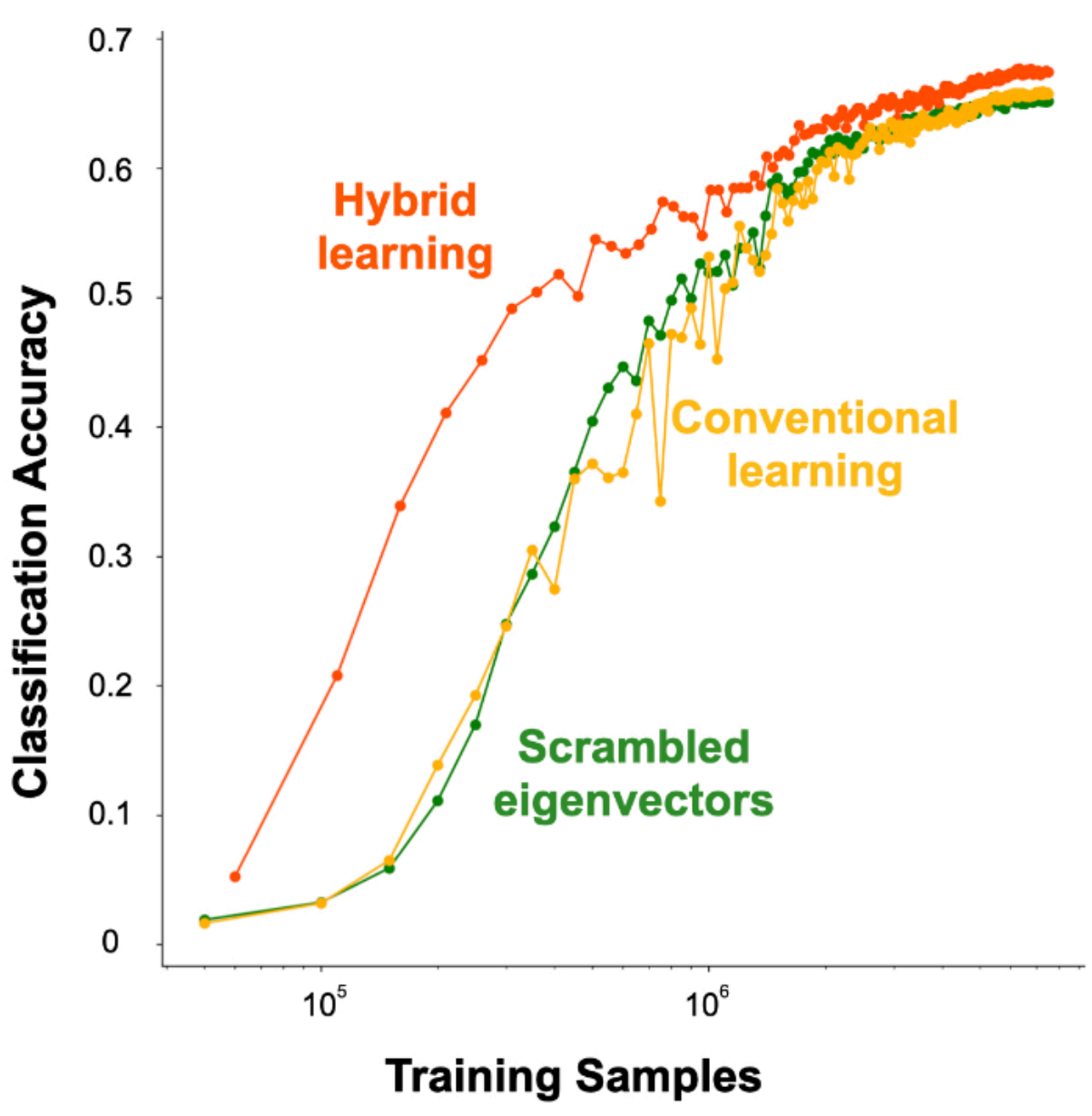

**Figure S3: The structure of eigenvectors, not their distribution, drives improvements in classification accuracy.** To test whether performance gains afforded by unsupervised initialization arise from the specific structure of the PC eigenvectors, the unsupervised eigenbasis was taken and the elements of its eigenvectors were permuted. Both the original and permuted models were then trained on miniImageNet classification. For comparison, a standard randomly initialized model was also trained. Classification accuracy on miniImageNet is plotted as a function of training samples: models initialized with the intact eigenbasis (red) consistently achieve higher accuracy with fewer samples, whereas those with shuffled eigenvectors (green) require substantially more training data to reach comparable performance. The performance of models with shuffled eigenvectors is similar to that of a model trained from random initialization (yellow). These results indicate that the organization of eigenvectors, rather than merely the weight distributions, underlies the observed improvements in classification performance.

**Hybrid learning**

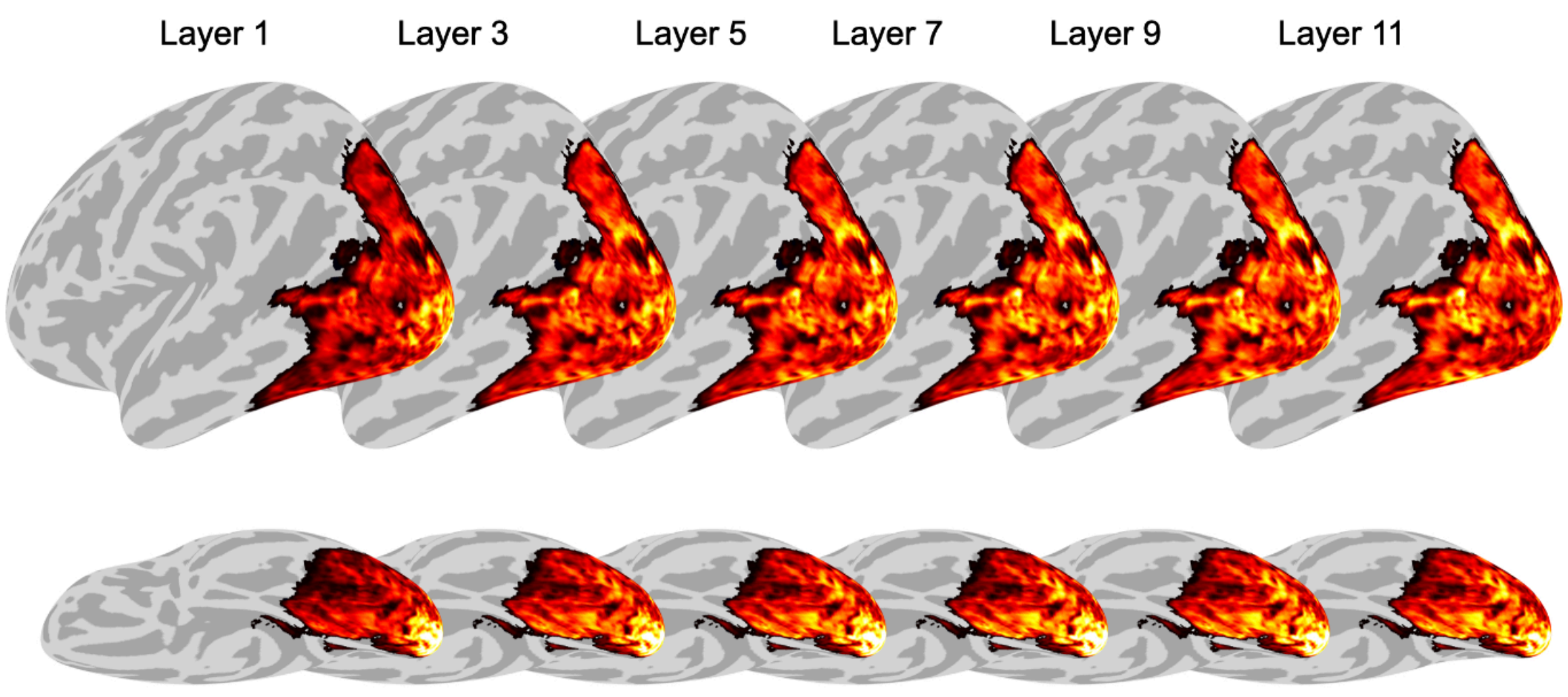

**Conventional learning**

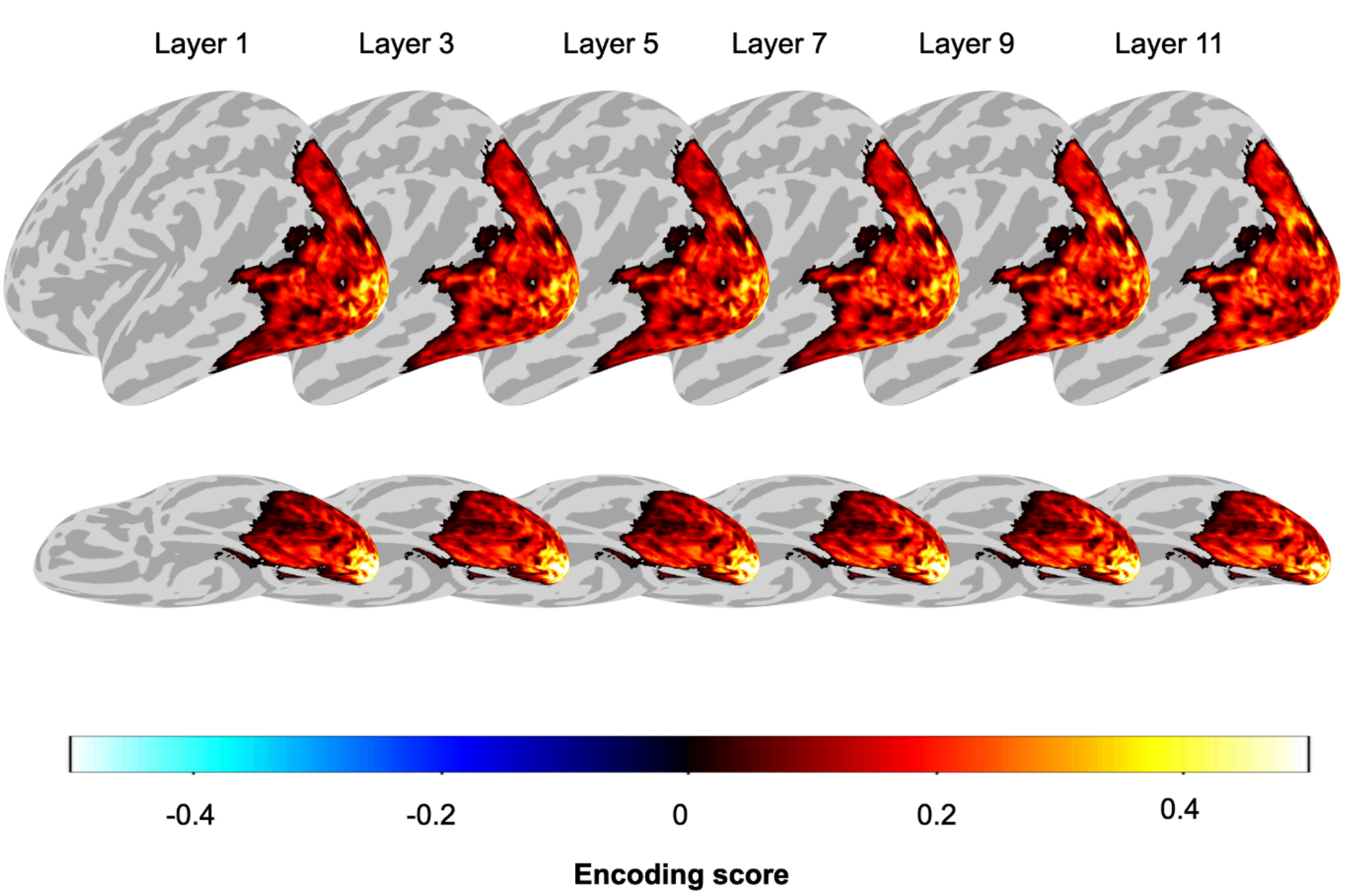

**Figure S4. Layer-wise encoding performance for Subject 0 comparing hybrid learning (top) versus conventional learning (bottom)**. Brain maps display encoding scores across successive layers for hybrid and conventional models trained on 1,000 images. Both models predict responses across visual cortex, but hybrid learning yields higher encoding scores in higher-level regions.

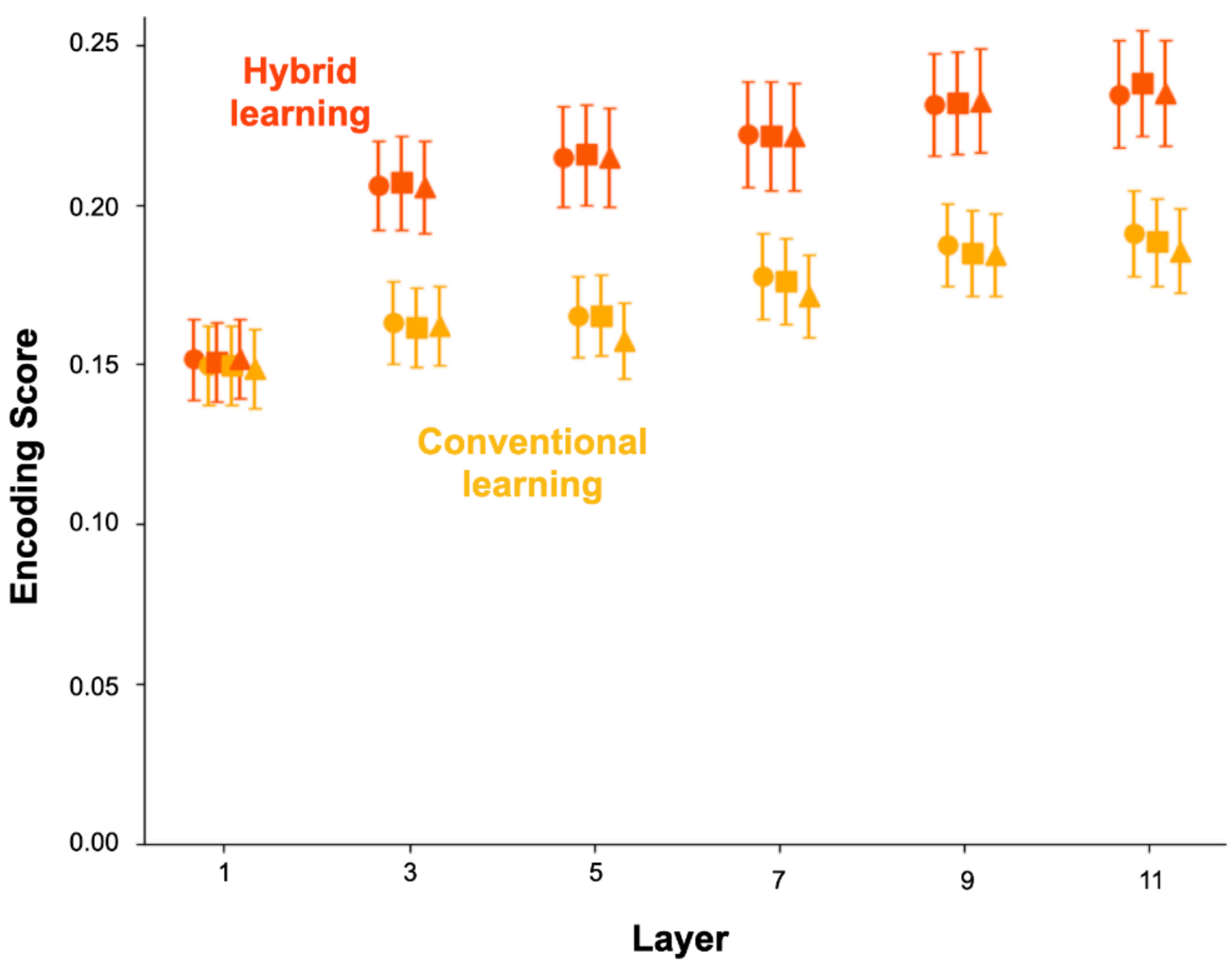

**Figure S5. Networks trained using hybrid and conventional learning exhibit consistent encoding performance with different random training sets.** Comparison of hybrid learning (red) versus conventional supervision (yellow) across network layers. Different markers represent distinct training image sets; error bars reflect cross-subject variability. Hybrid learning yields consistently higher encoding scores than conventional learning in deeper layers, and this effect is not contingent on the specific images selected for training.